\documentclass[hidelinks]{isprs}
\usepackage{subfigure}
\usepackage{import}
\usepackage{setspace}
\usepackage{geometry} 
\usepackage{epstopdf}
\usepackage[labelsep=period]{caption}  

\usepackage{hyperref}
\usepackage{graphicx}
\usepackage[export]{adjustbox}
\usepackage{amsmath,amsfonts,amssymb}
\usepackage[acronyms]{glossaries}
\usepackage[sort]{natbib} 
	\renewcommand{\refname}{REFERENCES} 
\usepackage{color, soul}
\usepackage{diagbox}	
\usepackage{enumitem}
\usepackage{gensymb}
\usepackage[nameinlink, capitalize]{cleveref}
\usepackage{hhline}
\usepackage{inputenc}
\usepackage{todonotes}

\usepackage[percent]{overpic}

\usepackage{xspace}

\newcommand{\ie}{i.e.\ }

\newcommand{\eg}{e.g.\ }

\newcommand{\cf}{cf.\ }

\newcommand{\wrt}{w.r.t.\ }

\newcommand{\approxi}{approx.\ }

\newcommand{\m}{\,m\xspace}

\newcommand{\fps}{\,fps\xspace}

\newcommand{\percent}{\,\%\xspace}

\newcommand{\KITTI}{KITTI\xspace}

\newcommand{\COLMAP}{COLMAP\xspace}

\newcommand{\ETHThreeD}{$\text{ETH3D}$\xspace}

\newcommand{\snSGM}{SGM$^{sn}$\xspace}

\begin{document}

\title{Self-Supervised Learning for Monocular Depth Estimation \protect\\ from Aerial Imagery}

\author{
M. Hermann\ \textsuperscript{a,b,c}, B. Ruf\ \textsuperscript{a,b,c}\thanks{Corresponding author}\ \ , M. Weinmann\ \textsuperscript{b}, S. Hinz\ \textsuperscript{b}
}

\address{
\textsuperscript{a}Fraunhofer IOSB, Karlsruhe, Germany -\\ \{max.hermann, boitumelo.ruf\}@iosb.fraunhofer.de\\
\textsuperscript{b}Institute of Photogrammetry and Remote Sensing, KIT, Karlsruhe, Germany -\\ \{max.hermann, boitumelo.ruf, martin.weinmann, stefan.hinz\}@kit.edu \\
\textsuperscript{c}Fraunhofer Center for Machine Learning
}


\commission{II, }{II} 
\workinggroup{4, } 
\icwg{I/II}   


\newacronym{ASG}{ASG}{{Average Shading Gradient}}

\newacronym[shortplural={CNNs}, firstplural={convolutional neural networks (CNNs)}, longplural={convolutional neural networks}]{CNN}{CNN}{convolutional neural network}
\newacronym{COTS}{COTS}{commercial off-the-shelf}
\newacronym[shortplural={CRFs}, longplural={{Conditional Random Fields}}]{CRF}{CRF}{{Conditional Random Field}}
\newacronym{CT}{CT}{{Census Transform}}

\newacronym{DLT}{DLT}{{Direct Linear Transformation}}
\newacronym{DoG}{DoG}{{Difference of Gaussian}}

\newacronym{EPnP}{EPnP}{{Efficient Perspective-n-Point}}

\newacronym{GPGPU}{GPGPU}{general purpose computation on a {GPU}}
\newacronym{GPS}{GPS}{{Global Positioning System}}
\newacronym{GTA}{GTA V}{Grand Theft Auto V}

\newacronym{ICP}{ICP}{{Iterative-Closest-Point}}
\newacronym{IMU}{IMU}{{Inertial Measurement Unit}}
\newacronym{INS}{INS}{{Inertial Navigation System}}

\newacronym{LIDAR}{LiDAR}{{Light Detection and Ranging}}
\newacronym{L1-rel}{$\text{L1-rel}$}{{relative $\text{L1}$-Norm}}
\newacronym{L1-abs}{$\text{L1-abs}$}{{absolute $\text{L1}$-Norm}}

\newacronym[shortplural={MRFs}, longplural={{Markov Random Fields}}]{MRF}{MRF}{{Markov Random Field}}

\newacronym{MVS}{MVS}{Multi-View Stereo}

\newacronym{NCC}{NCC}{normalized cross correlation}

\newacronym{PCL}{PCL}{{Point Cloud Library}}

\newacronym{RANSAC}{RANSAC}{{Random Sampling Consensus}}
\newacronym{RMSE}{RMSE}{{Root Mean Square Error}}
\newacronym[shortplural={ROIs}, longplural={regions of interest}]{ROI}{RoI}{region of interest}

\newacronym{SAD}{SAD}{sum of absolute differences}
\newacronym{SFM}{SfM}{{Structure-from-Motion}}
\newacronym{SGBM}{SGBM}{{Semi-Global Block Matching}}
\newacronym{SGM}{SGM}{{Semi-Global Matching}}
\newacronym{SMDE}{SMDE}{{Self-supervised Monocular Depth Estimation}}
\newacronym{SSIM}{SSIM}{Structural Similarity}
\newacronym[shortplural={STNs}, longplural={Spatial Transformer Networks}]{STN}{STN}{{Spatial Transformer Network}}

\newacronym[shortplural={UAVs}, longplural={unmanned aerial vehicles}]{UAV}{UAV}{unmanned aerial vehicle}

\newacronym{WTA}{WTA}{winner-takes-it-all}

\abstract{
Supervised learning based methods for monocular depth estimation usually require large amounts of extensively annotated training data. %
In the case of aerial imagery, this ground truth is particularly difficult to acquire. %
Therefore, in this paper, we present a method for self-supervised learning for monocular depth estimation from aerial imagery that does not require annotated training data. %
For this, we only use an image sequence from a single moving camera and learn to simultaneously estimate depth and pose information. %
By sharing the weights between pose and depth estimation, we achieve a relatively small model, which favors real-time application. %
We evaluate our approach on three diverse datasets and compare the results to conventional methods that estimate depth maps based on multi-view geometry. %
We achieve an accuracy $\delta_{1.25}$ of up to 93.5\,\%. %
In addition, we have paid particular attention to the generalization of a trained model to unknown data and the self-improving capabilities of our approach. %
We conclude that, even though the results of monocular depth estimation are inferior to those achieved by conventional methods, they are well suited to provide a good initialization for methods that rely on image matching or to provide estimates in regions where image matching fails, \eg occluded or texture-less regions. 
} 

\maketitle

\glsresetall 

\section{INTRODUCTION}
\label{sec:intro}

\sloppy

Dense depth estimation is one of the most important and intensively studied tasks in photogrammetric computer vision. %
It produces dense depth maps that contain depth estimates at each pixel and represent the 3D scene geometry from certain viewpoints. %
Depth maps are a key input to numerous applications, such as dense 3D reconstruction and model generation, navigation of autonomous vehicles such as robots, cars and \glspl*{UAV}, as well as scene interpretation and analysis. 
Given two or more images, that depict the same scene from different viewpoints, the process of dense depth estimation can be formulated as the problem of finding a dense correspondence field between the input images, which in turn can be transformed into a depth map, if the corresponding camera projection matrices are known. %
In this, it is assumed that the scene geometry does not change between the different images, which does not always hold, especially when trying to reconstruct a scene with dynamic objects, \eg an urban area with moving cars or pedestrians, by means of \gls*{SFM}.

\begin{figure}  
          \centering
          \includegraphics[width=\columnwidth]{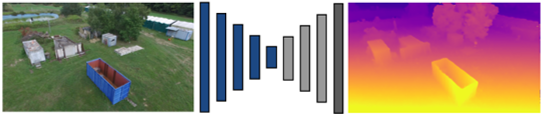} 
					\small
          \caption{
            Schematic illustration of deep learning based monocular depth estimation. %
              }
          \label{fig:image_first_page}%
    \end{figure}

Recent advancements in the field of deep learning have led to an increasing effort in attempting to learn how to hypothesize a depth image from a single input image \citep{eigen2014depth,Godard2016,li2015depth}. %
Figure \ref{fig:image_first_page} illustrates the general approach for this task. %
This process, known as \emph{Monocular Depth Estimation}, is motivated by the capabilities of humans to guess depth estimates from a single image of a known scene. %
Similar to the empirical knowledge that humans establish throughout their lifetime, state-of-the-art \glspl*{CNN} are able to efficiently learn discriminative image cues that allow them to infer depth information from a new, so far unseen, image. %
Evidently, this only holds if the scene depicted in the new image is the same or at least similar to the scene that is covered by the training data. %

There are a number of advantages to using monocular depth estimation instead of conventional image matching. %
For one, \gls*{SFM} is unstable when the camera is moving in the direction of the optical axis or if the camera has a narrow field-of-view. %
This is also the reason why image-based depth estimation in the context of autonomous driving is solely relying on a stereo camera setup. %
However, since the maximum depth range depends on the baseline between the images, a stereo camera is not always an option.
Furthermore, monocular depth estimation does not suffer from occlusions, which are apparent to image matching in urban environments or when using \glspl*{UAV} flying at low altitudes. %
Finally, one major advantage is the processing speed which can be achieved by monocular methods. %
Since they only require a single image to estimate the scene depth, they can operate at a much higher frequency than \gls*{SFM}-based methods, which need at least two images. %

\clearpage
So far, a majority of studies \citep{eigen2014depth,li2015depth,laina2016deeper} have focused on a supervised training of \glspl*{CNN} for the task of monocular depth estimation. %
They show that, with appropriate training data, state-of-the-art networks can even outperform conventional methods that rely on image matching. 
However, in many cases, supervised training is not feasible, first and foremost because it requires an appropriate dataset that contains labeled ground truth data. %
In the case of monocular depth estimation, this would mean to have a dataset with ground truth depth maps for each input image that is used for training. %
Acquiring such a dataset is cumbersome and costly. %
Thus, to the best of our knowledge, there only exist a few datasets \citep{Geiger2012,Schoeps2017eth3d, Silberman2012nyu} that provide suitable data to allow for a supervised learning for monocular depth estimation. %
However, none of them is appropriate to learn to predict a depth map from a single aerial image, since they all consist of terrestrial imagery.%

To overcome the limitation of such datasets, recent work on \emph{\gls*{SMDE}} has proven that it is also possible to train a network which is designed for monocular depth estimation by only using image pairs from a stereo camera or a video from a single moving camera. 
During self-supervised training, depth estimation is posed as a view synthesis and image reconstruction problem in which one image is transformed into the view of another camera given the predicted depth and matched with the real image of the second camera. %
A network has learned to predict the depth correctly, if the transformed image corresponds to the actual image taken by the second camera. %
The key advantage of the \gls*{SMDE} approach is that it does not require any special training data, and thus eliminates a major impediment of deep learning based approaches. %
This allows \gls*{SMDE} to be applied to any domain without the need for a costly acquisition of an appropriate training dataset. %

Based on the results and advancements achieved by \gls*{SMDE} in the context of autonomous driving and driver assistance, we adopt the approach in our work to learn monocular depth estimation from aerial imagery captured by a single camera mounted to a \acrlong*{COTS} \gls*{UAV}. %
Thus, in this work
\begin{itemize}
\item we show that it is possible to train a network in a self-supervised manner to predict a depth map from a single aerial image, not relying on any labeled ground truth during training, %
\item we compare the results achieved by our \gls*{SMDE} approach to results obtained by conventional methods that estimate depth maps based on multi-view geometry, and we argue why it can be feasible to rely on \gls*{SMDE} instead, %
\item we evaluate the overall performance of the trained model and draw a conclusion about how well it can be generalized to unknown scenes. %
\end{itemize}

We briefly review different approaches on supervised and self-supervised monocular depth estimation in \Cref{sec:related_work}. %
This is followed by a presentation of our methodology in \Cref{sec:methodology}. %
In this, we give a detailed description on the self-supervised training for monocular depth estimation. %
We evaluate our approach on three different datasets and present the results in \Cref{sec:eval}, together with a discussion, an ablation study and a comparison to traditional methods.
Furthermore, we provide a discussion on generalization and the self-improving capabilities of the presented approach. %
Finally, we provide a brief summary, concluding remarks and a short outlook on future work in \Cref{sec:conclusion}.

\section{RELATED WORK}%
\label{sec:related_work}
\sloppy

Approaches that allow for dense depth estimation from aerial imagery often rely on \glspl*{MRF}. %
In this, an energy functional is formulated and minimized to compute a smooth dense depth map while preserving depth continuities and geometric relationships, such as occlusions. %
In particular, the so-called \gls*{SGM} \citep{Hirschmueller2008}, which minimizes an energy functional by employing dynamic programming, has proven to be well suited for dense image matching from aerial imagery, due to its trade-off between accuracy and efficiency \citep{Rothermel2012, Wenzel2013Sure, Ruf2019efficient}. %
\gls{SGM} can be applied to images from a stereo camera setup as well as to images acquired from a single moving camera. %
If the latter is the case, information about the movement of the camera is required. 

\subsection{Supervised learning}
Besides classic approaches, there are also learning-based approaches to dense depth estimation. %
Most of them are based on \glspl*{CNN}, which are trained in a supervised manner using ground truth depth maps \citep{eigen2014depth,li2015depth,laina2016deeper}. %
However, gathering sufficient amounts of training data is costly, so the number of available datasets is small.
Especially in case of aerial imagery, we are not aware of any dataset that contains enough color images and dense depth maps to perform supervised training on a large scale. %
Even though some studies suggest to use synthetic datasets for training \citep{mayer2018makes,johnson2016driving}, it is still very time-consuming to generate large amounts of realistic data. %
Moreover, the use of a trained model across different domains, \eg transferring between terrestrial and aerial imagery, or training a model on a synthetic dataset and using it to real-world data, is still an open issue and doesn't always provide satisfying results. %

\subsection{Self-supervised learning}
A possible solution to not being dependent on a dataset with ground truth depth maps are self-supervised techniques. %
Instead of comparing the predicted depth map with a ground truth, the problem is modeled as the problem of novel view synthesis. %
In earlier works, this approach was used to generate new viewpoints from given imagery by estimating the depth of the scene \citep{flynn2016deepstereo,Xie2016}. %
In this, the depth map is only used as an auxiliary resource, since an understanding of the scene geometry is necessary to sample correct images. %
But there are also approaches that take advantage of this in order to predict depth maps. %
Here, the process of view synthesis is only used as a training method by comparing the synthetic image with a given image from the corresponding viewpoint. %
This is possible either by estimating the depth from several images as shown by \citet{Khot2019}, or by estimating the depth from only one image. %
For the process of sampling synthetic images, the extrinsic parameters between individual input images as well as the intrinsic parameters of the cameras must be known. %
This is why methods based on stereo cameras like the one proposed by \citet{Godard2016} are particularly suitable, since rotation and translation between left and right images are fixed and can be calibrated a priori. %
However, relying on a stereo camera setup is not always feasible. 
Consequently, there are methods that exploit the \gls*{SFM} paradigm during training and rely on images from only one camera by additionally estimating the relative transformation between the images in order to estimate the depth \citep{Zhou_2017_CVPR,Wang2017,mahjourian2018unsupervised}. %
This allows to learn directly from a video without the need for additional ground truth, making it much easier to obtain suitable training material. %
It is especially suitable for monocular depth estimation from aerial imagery captured by \acrlong*{COTS} \glspl*{UAV}. %

\subsection{Self-supervised learning from aerial imagery}
Most of the techniques described above derive from the context of autonomous driving. %
Aerial imagery, however, is very different from street recordings, since not only the flight altitude varies, but also the angles and  movement of the camera. %
\citet{knoebelreiter2018} show that a self-supervised approach can also be applied to aerial photographs. %
They rely on rectified stereo image pairs and only use the predicted depth maps as an initialization to a hybrid approach based on a \gls*{CNN} and \glspl*{CRF} \citep{knobelreiter2017end}. %

In contrast, we directly evaluate the predicted depth maps of our \acrlong*{SMDE} and do not rely on a secondary procedure to refine the results. %
Furthermore, we only use the pictures of one camera as training material. %
The training material is obtained from videos captured by a \gls*{UAV}. %
Our approach is somewhat comparable to the ones presented in the context of autonomous driving, since both depth and camera movements are estimated by our model.  %

\section{METHODOLOGY}%
\label{sec:methodology}
\sloppy

In the following, we present and discuss our methodology for \acrlong*{SMDE}\footnote{ Code available at: \url{https://github.com/Max-Hermann/SelfSupervisedAerialDepthEstimator/}}.
Since we use a learning-based approach, we have structured our methodology with respect to training and inference, whereby we have additionally divided the section on the training phase into four subsections referring to depth and pose estimation as well as image projection and image reconstruction, \ie loss calculation. %

\subsection{Training}
\label{sec:meth_training}

The training data consists of image sequences depicting a scene for which the network is to be trained. %
From these input images, triplets are formed which serve as an input to the training process. %
Each triplet should show a static scene from three different viewpoints. %
In this, it is crucial that there exists enough parallax between the images to predict the depth, but not so much that it would hinder a matching between the images. In order to fulfill these conditions, the individual images of a triplet are sampled from the video with a certain offset. %
This offset is empirically determined for each dataset and depends, among other factors, on the flight speed and the frame rate of the video. %

In the following, the middle image of a triplet is referred to as the reference image $I_\mathrm{ref}$ and the other two as matching images $I_\mathrm{left}, I_\mathrm{right}$. %
In addition, the intrinsic camera parameters are required for the training procedure. %

\begin{figure}[ht]
    \centering
    \def\svgwidth{\columnwidth}
    \input{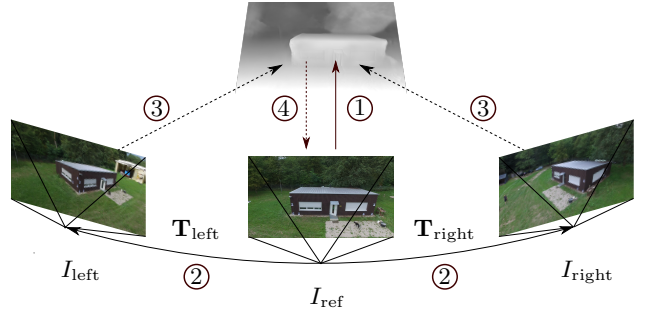}
    \small
    \caption{Training procedure with the estimation of depth (1) and pose (2) information, as well as the sampling of the synthetic images (3) and the calculation of the reconstruction error (4).}
    \label{fig:training_steps}
\end{figure} 

As shown in \Cref{fig:training_steps}, the training process is divided into four consecutive steps that are executed in each training iteration. %
1)~First, we perform monocular depth estimation to predict a depth map $D$ from $I_\mathrm{ref}$. %
2)~Afterwards, the relative rotation and translation between the reference image $I_\mathrm{ref}$ and the two adjacent images $I_\mathrm{left}, I_\mathrm{right}$ is estimated. %
3)~From these additional viewpoints, synthetic versions of the reference image are then sampled using the estimated depth map $D$ and relative camera poses $\mathbf{T}_\mathrm{k} = \left[\mathbf{R}_k\ \ \mathrm{t}_k \right]$. %
4)~In the last step, an image reconstruction error is calculated from these synthetic images and the original image. %
This error is backpropagated through the network, serving as the training loss.

Evidently, the depth and pose estimates will be of low quality at the beginning of the training process, resulting in a poor image reconstruction. %
With each training iteration, the network will improve its capability to learn how to predict the corresponding information by minimizing the training loss. %
It is assumed, that the network has learned to predict the depth map and relative camera poses correctly, if the synthetic images generated from $I_\mathrm{left}, I_\mathrm{right}$ correspond to $I_\mathrm{ref}$, \ie when the reconstruction error is at its minimum.

\subsubsection{Monocular depth estimation}
\label{sec:meth_training_depth}

For monocular depth estimation, we use a network topology that is based on the U-Net architecture developed by \citet{ronneberger2015u}. %
It consists of an encoder and a decoder connected with multiple skip connections to include both high-level and low-level features. %
Based on our ablation study (\cf \cref{sec:eval_ablation}), the ResNet18 architecture \citep{He2015} has shown to provide the best results when used as the encoder of the U-Net. %
As decoder, we adopt the approach of \citet{Godard2016}, which uses nearest neighbor upsampling. 
In contrast to a transposed convolution (deconvolution), it does not require any additional parameters that need to be learned during training. %
The last layer of the \mbox{U-Net} uses a \emph{Sigmoid} activation function to predict a depth map with the same dimensions as the input image. %

\subsubsection{Pose estimation}
\label{sec:meth_training_pose}

\begin{figure}[ht]
    \centering
    \def\svgwidth{0.9\columnwidth}
    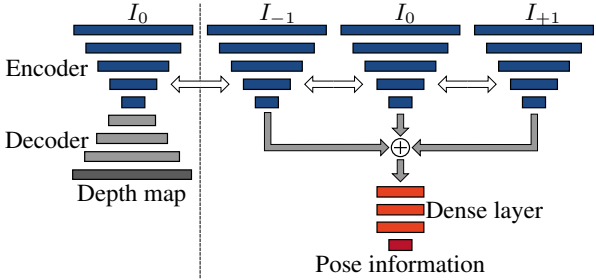
    \small
    \caption{
		Network architecture with depth estimation (left) and pose estimation (right) networks, sharing the encoder (blue). %
		For pose estimation, the feature maps are concatenated and passed through the dense layers with a final global average pooling. %
		}
    \label{fig:net_work_architecture}
\end{figure}

Because we focus on real-time computation, a small size of our resulting model is important to us. %
To reduce the number of trainable parameters, which in turn increases the training speed, we share the weights of our depth and pose encoders. %
This means, that our network for pose estimation uses the same layers of the U-Net encoder used for depth estimation to compute feature maps from the input images. %
We feed all three images through the shared encoder and concatenate the resulting feature maps. %
These are then passed through three dense layers with subsequent global average pooling. %
In this way, the parameters of the relative rotation $\mathbf{R}$ and translation $\mathrm{t}$ with respect to the reference image are estimated for each matching image. %
This approach is similar to the architecture shown in \citep{Zhou_2017_CVPR}. %
To sum up, although we differentiate between the networks for depth and pose estimation, both share a common encoder as depicted in \Cref{fig:net_work_architecture}. %

\subsubsection{Image projection}
\label{sec:meth_training_projection}

To learn the complex process of image projection and image sampling, and thus synthetic view generation, we use the concept of the \emph{\glspl*{STN}} presented by \citet{Jaderberg2015}. %
\glspl*{STN} allow to learn only the parameters which are necessary for sampling instead of learning the whole algorithm. %
As sampling method, we use bilinear interpolation to generate synthetic images from a three-dimensional grid constructed from the predicted depth maps. %
The required parameters are formalized below:
\begin{equation}
    I_{\mathrm{ k} \to \mathrm{ref }}=I_\mathrm{ k }\left[ proj(D ,\mathbf{T}_\mathrm{k} ,\mathbf{K} )\right]
\end{equation}
The projection $ I_{\mathrm{ k} \to \mathrm{ref }} $ of the source image $ I_\mathrm{k} $ onto the reference image $ I_\mathrm{ref} $ is formed by the sampling operator $ [\cdot] $ from the pixel coordinates. %
In this context, $ D $ is the predicted depth map of the reference image, $\mathbf{T}_\mathrm{k} $ is the transformation matrix into the camera coordinate system of the source image provided by the pose network and $ \mathbf{K} $ is the intrinsic camera matrix. %

For a more stable training, we use a multiscale estimation by outputting a depth map after each upsampling layer. %
To compare the resulting image projections with the reference image, the resolution of the depth map is adjusted. %
Instead of downsampling the source image to the resolution of the depth map, we use the approach presented by \citet{Godard2018} and upsample the depth maps to the resolution of the source image. 

\subsubsection{Image reconstruction and loss calculation}
\label{sec:meth_training_reconstruction}

In contrast to supervised learning methods, the self-supervised approach does not rely on a ground truth to which a loss function can be formulated in order to train the network. %
Thus, instead of comparing the predicted depth map with a ground truth depth map, we train the network to reconstruct the reference image from the matching images, given the predicted depth and relative transformation between the cameras. %
Accordingly, our training loss only indirectly allows reasoning about the quality of the depth maps, since it only measures the quality of the reconstructed images. %

However, due to the structure of the networks, a good image reconstruction is only possible if depth and pose estimation provide accurate values. %
Therefore, the loss function is modeled to evaluate the quality of the image reconstruction. %
We additionally adjust the loss function to enforce smooth depth maps, which helps in weakly textured image regions. %
Thus, the aggregated loss $L$ is calculated as follows:
\begin{equation}
    L=L_{ p }+\lambda L_{ s } ,
\end{equation}
with $L_{p}$ being the photometric loss and $L_{s}$ being the smoothing loss weighted with the parameter $\lambda$. %
We empirically set the weighting factor to $\lambda = 0.001$. %

In order to account for occlusions, we use the pixel-wise minimum of the error maps from both matching images, since it can be assumed that areas in $I_\mathrm{ref}$ which are occluded in $I_\mathrm{left}$ will be visible in $I_\mathrm{right}$, and vica versa \citep{Kang2001}. %
Therefore, the photometric loss $L_{p}$ is composed of the minimum image reconstruction error $pe$ between the reference image $I_\mathrm{ref}$ and the two sampled matching images $ I_{k\to\mathrm{ref}} $: %
\begin{equation}
    L_{ p }=\min_{I_k}pe(I_\mathrm{ref}, I_{k\to\mathrm{ref}})
\end{equation}

Similar to \citep{Godard2016,Zhao2017,mahjourian2018unsupervised}, we employ a combined photometric loss consisting of a \emph{\gls{SSIM}} and a $\text{L1}$ loss: %
\begin{equation}
    pe(I_{ a },I_{ b })=\alpha\frac{ 1-SSIM\left( I_{ a },I_{ b } \right) }{ 2 }+(1-\alpha)\parallel I_{ a },I_{ b }\parallel_{ 1 } 
\end{equation}
We use a value of $\alpha = 0.15$, as this leads to a more stable convergence of our training.

To enforce smooth depth maps, we use an edge-aware smoothness loss like shown in \citep{Wang2017}. %
The underlying idea is that discontinuities in depth are accompanied by a change in the color gradient of the reference image. %
Because we encountered problems with degrading depth maps, we used the idea of depth normalization as presented in \citep{Wang2017}:
\begin{equation}
    L_{ s }=\left\lvert \partial_{ x } d /\overline{d}   \right\rvert e^{ -\left\lvert \partial_{ x } I_\mathrm{ref}  \right\rvert }+\left\lvert \partial_{ y } d /\overline{d}   \right\rvert e^{ -\left\lvert \partial_{ y } I_\mathrm{ref}  \right\rvert }
\end{equation}
In this context, the gradient of the normalized depth map $d /\overline{d}$ is weighted with the color gradient of $I_\mathrm{ref}$ in $x$~and~$y$ direction. 

\subsection{Inference}
\label{sec:meth_inference}

Since our approach aims for monocular depth estimation, the actual inference is performed on only one image. %
An arrangement of the images in groups of three is therefore no longer necessary here. %
Since image reconstruction using depth and pose information is only necessary for training, the network for estimating rotation and translation is no longer needed as well. %
Thus, during inference, we only use the network to predict the depth map from step one of our training procedure.
This leads to a significant reduction in the parameters of the resulting model and, in turn, to an increase of the inference speed. %

\section{EVALUATION}
\label{sec:eval}
\sloppy

In the following, we present and discuss the results achieved by our approach. %
First, we address the datasets and hyperparameters used. %
Then, we present the results achieved and compare them with results achieved by conventional methods. %
Additionally, we discuss the findings of our ablation study, done to evaluate different configurations of our model. %

\subsection{Datasets}
\label{sec:eval_datasets}

\begin{figure*}[ht]  
    \centering   
    \subfigure{
	    \rotatebox{90}{\scriptsize~Ground truth~~~~Pred. depth~~~~~~~~~~Input}
    	\includegraphics[width=0.98\textwidth]{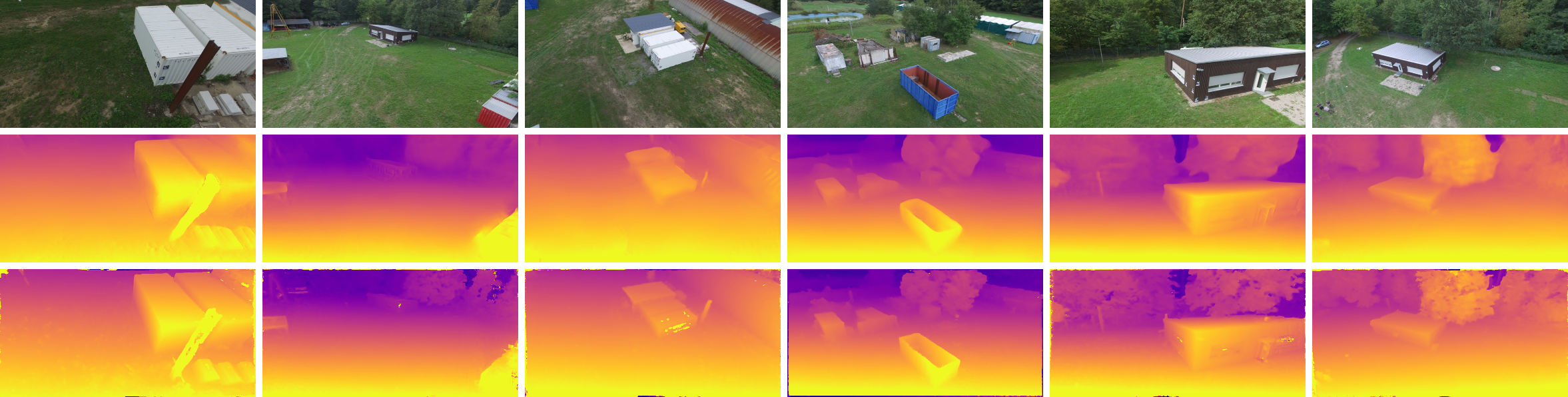}
    }
    \subfigure{  
	    \rotatebox{90}{\scriptsize~Ground truth~~~~Pred. depth~~~~~~~~~~Input}
    	\includegraphics[width=0.98\textwidth]{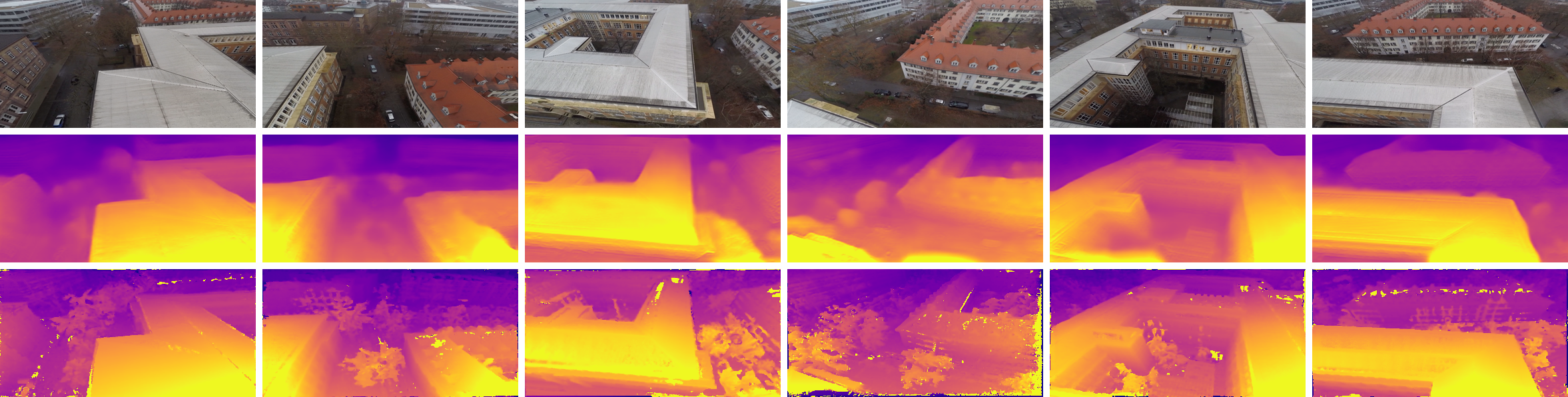}
    }
    \subfigure{
	    \rotatebox{90}{\scriptsize~Ground truth~~~~Pred. depth~~~~~~~~~~Input}
    	\includegraphics[width=0.98\textwidth]{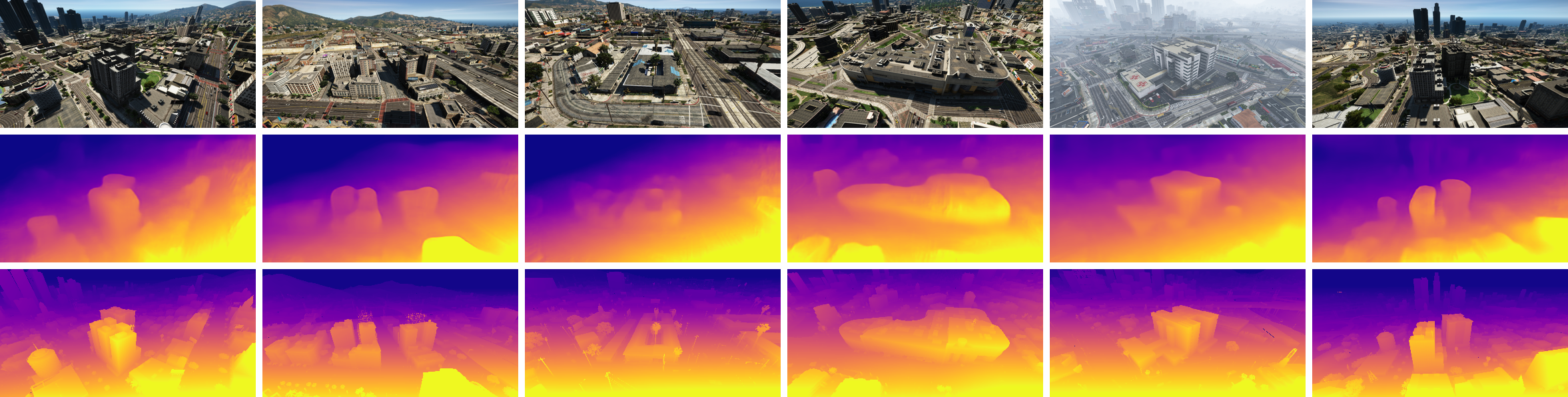}
	}
    \small
    \caption{
			Qualitative results  on all three datasets - from top to bottom: \emph{rural}, \emph{urban} and \emph{synthetic}. %
			The first row of each bundle shows the reference image for which the depth map is predicted. %
      The second row depicts the depth map that is predicted by the presented \gls*{SMDE} approach.
      In the last row, the corresponding ground truth depth map is shown. %
      In the case of the two real-world datasets, this is calculated by means of \COLMAP. %
    }	
    \label{fig:qualtitative_results}
\end{figure*}

In our experiments, we evaluate the overall performance of the \acrlong*{SMDE} approach on three different datasets: two private real-world datasets and one synthetic one (\cf \Cref{fig:qualtitative_results}). %
For the two real-world datasets, we generate ground truth data by employing state-of-the-art \gls*{SFM} methods provided in the software bundle \COLMAP \citep{Schoenberger2016mvs,Schoenberger2016sfm}. %
Since the third dataset is synthetically generated, we can rely on the simulation capabilities of modern rendering engines to extract the precise ground truth data. %

Our first dataset consists of video sequences showing three different rural scenes recorded from multiple different aerial viewpoints. %
In all sequences, the camera orbits around different objects of interest at different speeds, camera angles and altitudes. %
In this, the camera movement is only lateral. %
The second dataset contains multiple video sequences showing an urban scene. %
In contrast to the first dataset, the camera movement here is only forwards with sharp turns inducing high amounts of motion blur. %
Additionally, the image quality and lighting is reduced due to bad weather conditions such as rain or mist. %
Both the \emph{rural} and the \emph{urban} dataset were captured with a DJI Phantom 3 Professional. %
For our evaluation, we extracted 5,000 images from each of the first two datasets and processed them with \COLMAP to generate ground truth depth maps. %

Since \COLMAP does not produce perfectly accurate results, the ground truth depth maps still contain regions with errors. %
This causes apparent false positive and negative results that reduce the final score. %
To compensate for this issue, we synthetically generated a third dataset using the video game \gls*{GTA}. 
Synthetic data does not have the same level of detail, but it allows a manipulation of the environment. %
Thus, we simulated different kinds of flight trajectories, such as orbiting around buildings or flyovers with varying altitudes. %
Additionally, we randomized the weather conditions and time of day to create different scenes. %
To extract the depth maps from the video game, we used the software GTAVisionExport \citep{johnson2016driving} and adapted it to our needs. %
In this way, we have collected 5,000 color images with corresponding high-resolution depth maps. %
We truncated these depth maps at a maximum of $100$\m in order to provide comparable results to the other two datasets. %

We train and evaluate our model on all three datasets simultaneously. %
This leads to the same results as training three separate models, showing that the model is capable to deal with diverse scenes and camera motions at the same time. %

\subsection{Implementation details}
\label{sec:eval_implementation}

We implemented our model using the deep learning framework TensorFlow in version 1.15. %
To this end, we used a NVIDIA Titan X GPU for training and inference. %
We used the Adam optimizer with $\beta_{1}=0.9$, $\beta_{2}=0.999$ and a learning rate of $0.0002$. %
All results shown here are achieved by training with a batch size of 20. %
Smaller batch sizes can lead to unstable training, which is discussed in more detail in \Cref{sec:batch_size}. %
In total, our model contains 21 million parameters at training time. %
Since the network for pose estimation can be omitted for inference, the model then only has a size of 17 million parameters. %

The maximum training duration is difficult to determine, since training loss and evaluation error are decoupled. %
We have trained our networks from scratch up to a flattening of the training loss. %
By evaluation after each training period, we found out that the evaluation error does not improve after about one third of the total training time. %
We therefore approximate the maximum training time until the epoch that achieves the best result for the first time. %
With an image resolution of $384\times224$ pixels, our network needs about 100 epochs with a training duration of \approxi 24 hours for the full dataset consisting of 15,000 images. %
The training time decreases proportionally with a reduction of the image resolution. %

Additionally, we augmented the training data with randomly cropped and scaled copies of the input images. %
In order to avoid a bias in the camera movement, we flipped the images of a training triplet with a 50\,\% probability. %
This reverses the motion that is to be estimated by the pose estimation network. %
Finally, we adjusted contrast, brightness, saturation and hue with random percentage deviations of up to $0.2$, $0.2$, $0.2$ and $0.1$. %

Without optimizations, our model reaches speeds of more than 20\fps at a resolution of $384\times224$ pixels during inference. %
This includes the pre-processing of images such as resizing, as well as loading them into the memory of the GPU and saving them to the hard drive. %
To measure the processing speed, the model was exported as a frozen graph with a batch size of 100. %

\subsection{Experimental results}
\label{sec:eval_results}

We quantitatively assess the results of the \acrlong*{SMDE} approach with respect to the ground truth based on three metrics, namely the \gls*{RMSE}, the \gls*{L1-rel} as described in \citep{Ruf2019efficient} and the Accuracy. %
The latter is formulated in \citep{Godard2018} as: %
\begin{equation}
    \delta_{\theta}\left( d, \hat{d} \right) =\frac{ 1 }{ m }\sum^{m}_{i=1}{ \max\left(\frac{ d_{i} }{\hat{d}_{i}  }, \frac{ \hat{d}_{i} }{d_{i}}\right) < \theta}
\end{equation}
Originating within the context of image-based classification, it renders a pixel within the estimated depth map as correct, if the estimate is within a certain threshold $\theta$ to the corresponding measurement in the ground truth depth map. %
Thus, $\delta_{1.25}$ describes the fraction of pixels ($m$) of the ground truth depth map for which an estimate exists and for which the difference between the estimate $d$ and the ground truth $\hat{d}$ is not higher than $25$\percent of $\hat{d}$. %
This measurement is also used by the \KITTI \citep{Menze2015CVPR} and \ETHThreeD \citep{Schoeps2017eth3d} benchmarks. %

In our evaluation, the relative accuracy measures, \ie \gls*{L1-rel} and  $\delta_{\theta}$, are the most meaningful ones, since the different datasets differ greatly in the depicted scene depth and since the depth maps estimated by monocular depth estimation typically are free of a metric scale. %
To adjust the different resolution and depth range between an estimated depth map and the corresponding ground truth depth map, we have applied a simple image resizing and a median scaling. %
Furthermore, during evaluation, we use a $90/10$ split of training and evaluation data. %

\begin{table}[h] 
    \centering
    \begin{tabular*}{\columnwidth}{@{\extracolsep{\fill}}|l|c|c|c|c|}
        \hline
        \rule{0pt}{\normalbaselineskip}
        Dataset			&	\gls*{RMSE} & \gls*{L1-rel} &	$\delta_{1.25}$	&	$\delta_{1.05}$	\\
        \hline
        \rule{0pt}{\normalbaselineskip}     
        Rural				& 2.216				& 0.171					& 0.935						&	0.698	\\
        ~Urban				& 4.430				& 1.118					& 0.722						& 0.326	\\
        ~Synthetic		& 8.337				& 0.081					& 0.910						& 0.578	\\
        \hline
    \end{tabular*}
    \small
    \caption{
			Quantitative results achieved by our approach. %
			The values represent the mean score, averaged over all images in the corresponding dataset. %
			The last two columns represent the accuracy at a threshold of $25$\percent and $5$\percent \wrt to the ground truth. %
		}
    \label{tab:results_of_combined_model}
\end{table}

As shown in \Cref{tab:results_of_combined_model}, our results vary greatly between the different datasets. %
The best accuracy is achieved on the rural dataset, followed by the synthetic dataset. %
The results achieved on the urban dataset are considerably worse, which we assume to be caused by the bad picture quality and the faulty ground truth. %

\Cref{fig:qualtitative_results} shows examples of the predicted depth maps from each dataset, together with the corresponding reference image and the ground truth. %
In this, the depth is color coded, going from yellow (near) via red to blue (far). %
In the case of the two real-world datasets (rural and urban), the examples clearly show the errors in ground truth that have been induced by \COLMAP. %

Altogether, the results reveal that the presented self-supervised learning is capable of learning monocular depth estimation from aerial imagery. %
The quality of the resulting depth maps, in both quantitative and qualitative evaluation, is high with respect to the ground truth. %
Even though the model is incapable of predicting fine structures like vegetation and sharp edges, the essential scene geometry and objects are predicted correctly as can be derived from \Cref{fig:qualtitative_results}.



\subsection{Comparison to traditional methods}
\label{sec:eval_comparison}

In addition to an evaluation of the presented \acrlong*{SMDE} approach with respect to a ground truth, we have also compared the results to those achieved by conventional algorithms that are based on the \gls*{SGM} presented in \citep{Hirschmueller2008}. %
We have chosen \gls*{SGM}, since it is one of the most widely used algorithms for real-time depth estimation from two or more views. %
It allows to estimate accurate depth maps with reasonably low computational complexity. %

For our comparison, we have selected the \gls*{SGBM} variant provided in the computer vision library OpenCV, as well as the \snSGM extension presented in \citep{Ruf2019efficient}. %
The latter combines \gls*{SGM} with a multi-view plane-sweep sampling and incorporates local surface normals to improve the optimization. %
We have performed the comparison on the rural dataset, since it is the real-world dataset on which the best results are achieved by the \gls*{SMDE}. %
The results are listed in \Cref{tab:results_comparison}. %

\begin{table}[h] 
    \centering
    \begin{tabular*}{\columnwidth}{@{\extracolsep{\fill}}|l|c|c|c|c|}
        \hline
        \rule{0pt}{\normalbaselineskip}
        Method		 	& \gls*{L1-rel} &	$\delta_{1.25}$	&	$\delta_{1.15}$ 	&	$\delta_{1.05}$	\\
        \hline
        \rule{0pt}{\normalbaselineskip}     
        \gls*{SMDE}	& 0.171					& 0.935						&	0.894							&	0.698	\\
        ~\gls*{SGBM}				& 0.052					& 0.986						& 0.876							& 0.767 \\
        ~\snSGM			& 0.042					& 0.968						& 0.951							& 0.890 \\
        \hline
    \end{tabular*}
    \small
    \caption{
			Comparison of the presented \gls*{SMDE} approach to two variants of the \gls*{SGM} algorithm that perform depth estimation by epipolar geometry, namely \snSGM from \citep{Ruf2019efficient} and \gls*{SGBM} from OpenCV. %
		}
    \label{tab:results_comparison}
\end{table}

It is evident, that the results of the \gls*{SMDE} are inferior to those achieved by \gls*{SGM}. %
However, monocular depth estimation has a number of advantages compared to methods that rely on image matching. %
For one, it does not suffer from instabilities that are induced to \gls*{SFM}-based methods due to a bad choice of the camera movement. %
Furthermore, monocular methods produce fully dense depth maps, providing estimates in regions where image matching algorithms often fail, such as untextured or occluded areas.
And finally, since the prediction of the depth map only requires a single input image, such methods can operate at a much higher frequency as they do not need to wait until the camera has moved to a new viewpoint.

\subsection{Ablation study}
\label{sec:eval_ablation}

\paragraph*{Encoder topologies}

To study the impact of network topologies, we evaluated the use of three different architectures for the shared encoder as shown in \Cref{tab:archs_comparison}. %
All three topologies, namely VGG-Net16~\citep{Simonyan2014}, ResNet18~\citep{He2015} and DenseNet~\citep{Huang2016}, are suitable for learning meaningful depth maps.
However, ResNet18 achieves the best results, while simultaneously having less parameters than the other topologies and can thus be trained faster. %
We have also tried deeper variants like ResNet32 or ResNet50, but due to the limited graphics card memory, we had to reduce our batch size, resulting in very unstable training. 

\begin{table}[ht] 
    \centering
    \begin{tabular*}{\columnwidth}{@{\extracolsep{\fill}}|l|c|c|c|}
        \hline
        \rule{0pt}{\normalbaselineskip}
        Architecture		& \gls*{RMSE}				& \gls*{L1-rel}		& $\delta_{1.25}$	\\
        \hline
        \rule{0pt}{\normalbaselineskip}
        VGG$16$				& 2.406						& 1.985					&	\textbf{0.927}	\\
        ~ResNet$18$			& \textbf{2.191}	& \textbf{0.734}		& 0.926	\\
        ~DenseNet 			& 2.548					& 5.370						& 0.912 \\
        \hline
    \end{tabular*}
    \small
    \caption{
			Comparison of different network topologies used for the encoder. 
		}
    \label{tab:archs_comparison}
\end{table}

\paragraph*{Batch size}\label{sec:batch_size}
During our final training, we have used a batch size of $20$, which was found empirically. %
We observe very unstable training if we used smaller batch sizes. 
In particular during early iterations, the training tends to diverge and get stuck in a local minimum. %
We assume this problem arises from the pose estimation network, since datasets with a complex camera movement, \ie high amounts of rotation without forward motion, are especially affected. %
One possible explanation is that the encoder is shared between the depth estimation network and the pose estimation network. 
And since the pose estimation depends on depth information and vice versa, this could potentially create an endless loop in which the training does not converge. %
Using larger batch sizes increases the probability of processing different varieties of camera movements in one batch, and thus prevents divergence in early training iterations. %
After the network starts to predict reasonable depth maps, the batch size can be reduced. %
Furthermore, our experiments have shown, that batch sizes larger than $20$ do not tend to produce better results but restrict rather the use of a higher input resolution due to the confined memory on the GPU. %

\paragraph*{Input resolution}
The resolution of the input and output data has a great influence on the training, as it is, depending on the batch size used, limited by the memory of the graphics card. %
Our results shown in \Cref{tab:results_of_combined_model} are achieved with an image resolution of $384\times224$ pixels. %
However, our model still produces good results when trained with a very low resolution of $192\times96$ pixels. %
Using the higher resolution increases the training time by a factor of four, achieving a quantitative improvement of only 2\percent. %

\paragraph*{Size of image bundle}
Finally, to test whether our model benefits from more than three input images during training, we use sequences of five images, with four matching images projected onto the reference image. %
Our experiments show that this does not lead to superior results. %
We assume that the use of more matching images increases the probability of false positives in the process of image sampling, \ie pixels that have a high similarity but are sampled from a wrong location, which hinders the training procedure.

\subsection{Generalization and self-improving capabilities}
\label{sec:continual_learning}

\begin{figure}
      \centering
      \includegraphics[width=\columnwidth]{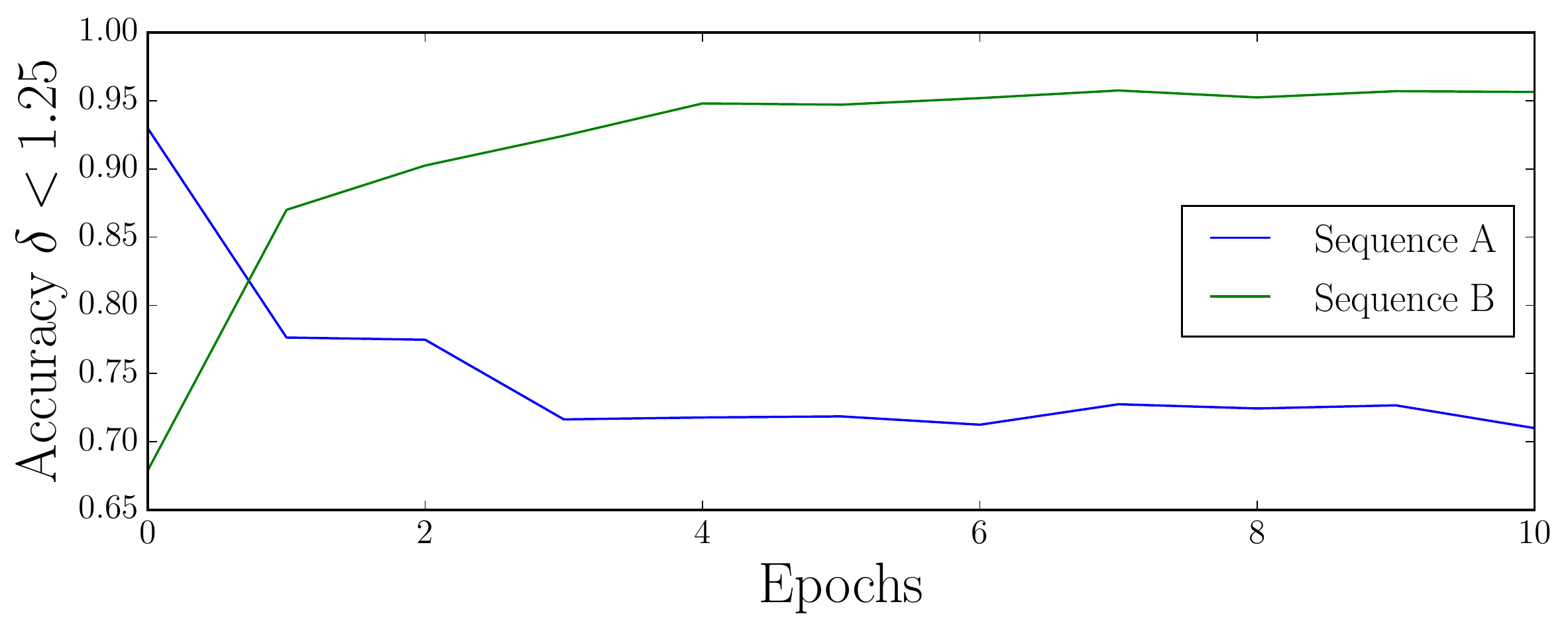} 
			\small
      \caption{
      	Evaluation of generalization and self-improving capabilities. %
      	After the network is trained until convergence on Sequence A, it is evaluated and fine-tuned on a new, yet similar Sequence B. %
      	The graph shows the accuracy achieved by the model on the two datasets during fine-tuning. %
      	}
      \label{fig:relearn_switch}%
\end{figure}

In order to evaluate the capabilities of our approach to generalize to unknown data, we use a setup that contains two short video sequences without any overlap or common objects, but similar conditions, in terms of camera angle and weather. %
We begin by training a base model until convergence on the first sequence (Sequence A) and then evaluate on the second one (Sequence B). %
As depicted in \Cref{fig:relearn_switch}, the accuracy $\delta_{1.25}$ drops by \approxi $25$\percent when switching to a new, yet similar sequence. %
From this, we conclude that the model generalizes rather moderately to unknown data, which hinders a practical use of the model. %
However, since it is a self-supervised method and does not require special ground truth data for training, the model can easily be retrained to learn the new sequence. %
This is often referred to as self-improving capabilities. %

In our experiments, we evaluate how fast the model can be fine-tuned to the new data. %
To do so, we have not fixed the weights of our resulting model but continued training the network with the new Sequence B.
As shown in \Cref{fig:relearn_switch}, the model can be fine-tuned within 2-4 epochs to a similar accuracy as initially achieved on Sequence A. %
One epoch corresponds to a training time of \approxi 2 minutes with the same configuration as stated in \Cref{sec:eval_implementation}. %
This fine-tuning is about 10 times faster than training from scratch, simultaneously achieving equal results. %

Yet, the experiments also show that the network tends to forget the old sequence, as the accuracy achieved on Sequence A decreased with each epoch in which the model is fine-tuned on Sequence B. %
To prevent this, we also employed in another experiment a combined fine-tuning with images of both sequences, leading to the same increase of the accuracy on Sequence B, while maintaining the accuracy on Sequence A. %
This, however, again increases the training time, since the dataset is bigger. %

Depending on the application, it should be decided whether it is more important to remember the old scene or to achieve a quick fine-tuning. %
In both cases, a base model that is as diverse as possible should be advantageous and can fine-tuned for a specialized use case. %
However, the method of self-improvement has also some disadvantages. %
In contrast to a standard application, in which only the depth estimation network is needed, the process of self-improvement also needs the network for pose estimation, increasing the size of the resulting model. %
Furthermore, not only the forward pass through the network is to be processed, as done during inference, but also the back-propagation in order to adjust the weights during the training, which increases the processing time for each input image.

\section{CONCLUSION \& FUTURE WORK}
\label{sec:conclusion}

\sloppy

To summarize, we show that it is possible to flexibly train a deep \acrlong*{CNN} to perform monocular depth estimation from aerial imagery. %
Our method relies on a self-supervised learning procedure that does not require any special training data and thus can be applied to any image sequences from a video that was captured from a static scene with a moving camera. %
During training, our networks learn to predict depth information from a single input image as well as relative camera poses between three images. %
During inference, our approach only requires the network for depth estimation, decreasing the size of the resulting model and the processing time. %
We reduce the amount of trainable parameters by sharing the weights between the different encoders and by using simple upscaling instead of deconvolution, which makes our model also suitable for the use on systems with reduced hardware capabilities. %

The conducted evaluation suggests a direct utilization of the estimated depth maps for a number of applications, such as real-time 3D modeling, navigation or scene interpretation without any further refinement. %
Even though the quality of the obtained depth maps is evidently inferior to the results achieved by conventional methods based on image matching, \eg \acrlong*{SGM}, monocular depth estimation can operate at a higher frequency and does not suffer from typical drawbacks of \gls*{SFM}, such as occlusions or instabilities associated with inappropriate camera trajectories. %
In conclusion, \acrlong*{SMDE} is well suited to complement conventional methods, for example by providing a good initialization or by providing estimates in regions where image matching fails, \eg occluded or texture-less regions. %

Furthermore, our experiments on practical use show that a trained model generalizes rather poorly and that an application of a trained model to a new dataset causes a significant drop in accuracy. %
However, by exploiting the self-improvement capability of our approach, we were able to show that a model can quickly be fine-tuned to the new data. %
Since we have only been using offline learning so far, an investigation on whether the self-improving capabilities of our model can also be realized with online learning is still to be done. %

Additionally, a generalization to changes applied to data which has been seen during training, such as different weather or lighting conditions, as well modifications in the geometry of familiar scenes, is to be evaluated. %
This is extremely relevant for the use case of repeated flyovers, needed for the application of change detection or for the extension and update of existing 3D models. %

Finally, the fact that the depth maps of monocular depth estimation methods do not have a metric scale makes it difficult to use them for accurate distance measures. %
Possible solutions to this could be the augmentation of the training data with a small number of metric depth maps or the use of a calibrated pose estimation method. %
Possible approaches would be the separation of depth and pose network, as well as the use of classical approaches of image processing, which would also help to handle more complex camera movements.
However, an integration of conventional methods would reduce the benefit of an end-to-end solution and the advantage of high processing rates. %

{\footnotesize 
	\begin{spacing}{0.7}
    \setlength{\bibsep}{2pt}
		\bibliography{self-supervised-depth_biblio} 
	\end{spacing}
}

\end{document}